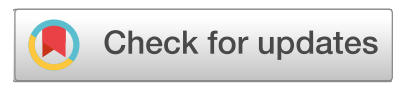

OPEN

# De-identification is not enough: a comparison between de-identified and synthetic clinical notes

Atiquer Rahman Sarkar[1]✉, Yao-Shun Chuang[2], Noman Mohammed[1] & Xiaoqian Jiang[2]

**For sharing privacy-sensitive data, de-identification is commonly regarded as adequate for safeguarding privacy. Synthetic data is also being considered as a privacy-preserving alternative. Recent successes with numerical and tabular data generative models and the breakthroughs in large generative language models raise the question of whether synthetically generated clinical notes could be a viable alternative to real notes for research purposes. In this work, we demonstrated that (i) de-identification of real clinical notes does not protect records against a membership inference attack, (ii) proposed a novel approach to generate synthetic clinical notes using the current state-of-the-art large language models, (iii) evaluated the performance of the synthetically generated notes in a clinical domain task, and (iv) proposed a way to mount a membership inference attack where the target model is trained with synthetic data. We observed that when synthetically generated notes closely match the performance of real data, they also exhibit similar privacy concerns to the real data. Whether other approaches to synthetically generated clinical notes could offer better trade-offs and become a better alternative to sensitive real notes warrants further investigation.**

Doctors write clinical notes for patients, which contain a wealth of information about the patient's history, reason for admission, care received, medications administered, and the patient's health progress from admission to discharge. These textual discharge notes are acknowledged for their rich information content[1]. Applying Natural Language Processing techniques to this untapped wealth of information could revolutionize the healthcare industry. For instance, by using the clinical notes dataset, researchers are working to develop various classification/prediction models such as to identify diseases based on admission state data, to predict mortality rates, life expectancy, readmission rates at the ICU, length of stay under care, etc.[2–5]. Applications like these can improve patient outcomes, optimize resource utilization, and streamline healthcare operations. Extracting information from clinical notes is a difficult and ongoing research area. However, there is limited participation in this research area largely due to the challenge of accessing relevant healthcare data[6]. The regulatory laws that govern the sharing and publishing of clinical notes, such as the Health Insurance Portability and Accountability Act (HIPAA) in the USA and the General Data Protection Regulation (GDPR) in the EU, rightfully restrict access to sensitive private data. HIPAA defines 18 types of data as protected health information (PHI) that must be redacted before data publication or sharing. Although de-identification may seem appealing, manual de-identification has been proven to be both expensive and prone to errors[7]. In contrast, automated de-identification mechanisms cannot yet detect all the protected health information (PHI) tokens[8,9]. Furthermore, some de-identification models that exhibited strong performance in specific datasets have been observed to perform inadequately in other datasets[7].

Researchers have mostly focused on the identity disclosure/re-identification risks of the clinical notes (e.g.,[6,10]). Recently, another type of privacy attack, the membership inference attack (MIA), has been introduced[11]. In some cases, disclosing the membership to a dataset/machine learning model's training set can become a very privacy-sensitive issue. Membership inference attack has been acknowledged as a serious breach of privacy (ref: NIST report[12]), and researchers have been continuously working to improve the success of the attack[13–15]. Consider a scenario where a machine learning classifier on a sensitive topic has been released, and this machine learning classifier was trained with de-identified clinical notes. Suppose that the nature of the classifier is so sensitive that disclosing membership information to its training dataset is damaging. Now, if an adversary somehow manages to get the doctor's note of a target person, the adversary can de-identify it and mount a membership inference attack on the released model and find out with reasonable probability whether the person's note was in the training dataset.

[1]Department of Computer Science, University of Manitoba, Winnipeg R3T 5V6, Canada. [2]McWilliams School of Biomedical Informatics, The University of Texas Health Science Center at Houston, Houston 77030, USA. ✉email: sarkarar@myumanitoba.ca

Another application of membership inference attacks is that they are widely recognized as the "*de facto standard*" for practically evaluating a model's privacy due to their straightforward implementation[15–17]. Consequently, the membership inference attack can be useful for identifying and measuring another kind of privacy vulnerability in machine learning models other than re-identification attacks. Although the membership inference attack has been previously mounted on models relying on structured health data[18], there are limited works that explored the MIA attacks on models that rely on textual clinical data. The privacy aspect of our work is unique from the existing works for bringing specific focus to de-identification and broader applicability (e.g., clinical note's usage is not restricted to only 'masked' or 'pre-trained' language models; more on this in the "Related works" section below). In this work, we demonstrated that de-identification of clinical notes does not protect a person's clinical note from the membership inference attack.

One potential solution to this privacy dilemma could be to generate and share synthetic clinical notes that preserve the information content of the original notes while respecting the regulatory requirements[19]. Consequently, it is necessary to evaluate the quality of the synthetic notes, both in terms of utility and privacy, to determine whether they can effectively replace the original clinical notes in real-world applications and research. In this work, we also proposed a synthetic clinical note-generation approach using any pre-trained generative large language model (LLM). We evaluated the utility and privacy of the synthetic note generation approach using GPT-3.5.

## Related works

Our privacy study is more aligned with the experimentation conducted by Vakili and Dalianis[20], but critical differences exist in the methodology and interpretation of the result. They concluded that the MIA attack fails to identify any privacy advantages gained by training the victim model with pseudonymized textual data. To point out the key differences with our privacy study, firstly, their proposed attack is restricted to pre-trained "masked" language models. More specifically, in their attack, they used what they refer to as "the normalized energy values," which depend on a set of masking patterns (ref. section 2.2[20]). Secondly, their clinical note usage significantly differs from our study in two aspects. Their masked model did not utilize the entire content of a MIMIC-III note. Their data points (i.e., notes under investigation) only comprised the sentences containing names. It is hard to imagine a medical application where only sentences containing names will be used. Lastly and most importantly, their perspective and conclusion are fundamentally different than ours. The authors claimed that since the MIA attack achieved similar success irrespective of whether a "real" or "pseudonymized" dataset is used, it *"fails to detect the privacy benefits of pseudonymizing data"*, and thus *"such attacks may be inadequate for evaluating token-level privacy preservation of PIIs"*. We agree with the validity of their conclusion, but we would like to point out that the question of whether MIA can evaluate token-level privacy (i.e., name alteration in this case) is not very pertinent regarding the objective of the MIA. The objective of the MIA attack is not to evaluate the token level privacy but rather to evaluate (infer) membership of the entire text content. We believe their conclusion should have been similar to ours, that pseudonymization does not protect notes from MIA attacks. In other words, the results from their study reinforce our findings although their study was focusing on a different aspect.

The work by Jagannatha et al.[21] performed MIA evaluation focusing very specifically on "pre-trained language models" (they experimented with GPT-2 and several variations of BERT). For GPT-2, the input comprises a token sequence, and the target is the next token in that sequence. For BERT and its variations, the input is the masked sequence, and the target is the masked tokens in those places. Our study is different in scope than their work as our study applies to any model that uses clinical notes for classification/prediction tasks. Moreover, if their adversary wishes to perform an MIA attack on a BERT-backed classification model, the adversary will require more intrusive access to the model than our adversary. In their black-box version, their adversary required access to the mean of the training error of the language model (ref. subsection 4.1.1, Jagannatha et al.[21]) and for the white-box version, they tried two variations: one required access to attention outputs from all the attention layers and the other required "*gradient values for each neural network layer by taking the squared-norm of all parameters in that layer*". Lastly, their study was not focused on de-identification and thus did not provide any conclusion or remarks on the impact of de-identification on MIA attacks. Nonetheless, their experimental result on MIMIC-III notes reinforces our findings that the MIA attack is possible despite de-identification.

The utility analyses of the synthetic clinical notes can be roughly categorized into three ways: (i) human evaluation (e.g.,[22]), (ii) linguistic property-based metrics, and (iii) downstream-task-based metric. Currently, there is no agreement on how to measure the quality of synthetic clinical notes, which makes it difficult to evaluate their effectiveness. The linguistic property-based metrics such as BLEU, ROUGE, METEOR, adapted Levenstein distance, adapted KL-divergence, BERT-score, etc., have been used with synthetic medical records[23–27]. These metrics focus more on the token and sentence level similarity. Though linguistic utility metrics capture some syntactic quality, they usually fail to adequately capture the semantic content of the clinical notes, as highlighted in previous research studies[22,28]. On the other hand, human evaluation is costly and not time-efficient. Therefore, downstream-task-based evaluation could be a proper choice for measuring the quality of the generated notes.

The downstream task-based evaluation compares the performance of the machine learning model trained with synthetic notes against models trained with real notes. Li et al.[29] used four language generation models, namely- CharRNN, SeqGAN, GPT-2, and conditional transformers, to generate synthetic notes. They measured the performance of the synthetic notes on a named entity recognition (NER) task. They generated 500 notes (max sequence length: 256 tokens) and used the BLEU score as their initial quality metric. They found that the GPT-2 model produced the best BLEU (BLEU 1-4) scores among those four models. For the rest of their experimentation, they used the GPT-2 generated synthetic notes when they needed synthetic data. Their study used three Named Entity Recognizer instances to compare the quality of the synthetic text to natural text. These three NER models were trained on (i) only natural data, (ii) only synthetic data, and (iii) natural+synthetic data. The trained models (i) and (ii) were evaluated against different annotated natural corpora and synthetic data. Model (iii) was evaluated against one natural corpora only (570 notes, as mentioned earlier). The results were interesting. It showed that the model trained on synthetic data performed better than the model trained on natural data in one case. They gave a possible explanation that the 500 synthetic notes which were used in training dataset contained more named entities than the natural corpora. Hence, the synthetic models were trained better. Also, the augmented (natural+synthetic) NER classifier performed better, which was expected as its training dataset was relatively larger.

Synthetic clinical text generation is a recent development. As pointed out by Li et al.[29], the previous works did not evaluate the utility of the notes even for common NLP tasks as basic as named entity recognition (NER). However, NER is not a core task in the healthcare sector, and this task cannot effectively measure the semantic or even syntactic quality of the synthetic text. On the other hand, the scarcity of research work in downstream-task-based evaluation is due to the fact that the usage of real clinical notes for a healthcare-related application is also scarce and has only begun to be investigated recently. Having discussed the various methods of evaluating synthetic clinical notes, we now turn to our contributions in this study to this field of research.

## Contributions

Until recently, GPT-2 was shown to outperform other text generative models such as CharRNN, SeqGAN, and conditional transformers[24,29]. There have been major breakthroughs in the large generative language model arena with the release of huge pre-trained models. For example, the jump in the number of parameters from GPT-2 to GPT-3 is from 1.5 billion to 175 billion. This work particularly focuses on the recent prompt-based language generative models GPT-3.5-turbo. OpenAI mentioned in their model's overview that GPT 3.5 is better than GPT-3. GPT 3.5 has been demonstrated to be able to generate human-like texts. For example, in simulated exams, GPT 3.5 is in the 87th percentile (score: 670/800) in SAT Evidence-Based Reading & Writing and in 54th percentile (score: 4/6) in Graduate Record Examination (GRE) Writing[30]. GPT-3.5-turbo has a maximum window length (prompt+completion) of 4097 tokens. In this work, we made four core contributions to this research area, which are as follows:

(1) *De-identification does not protect records from MIA attack* De-identification was previously considered a sufficient privacy-preserving step to make clinical notes available for public usage[7]. This study empirically demonstrated that de-identifying clinical notes fail to protect records from membership inference attacks. This finding has serious implications for cases where membership is a privacy-sensitive issue.

(2) *Clinical notes generation using an LLM* Our second contribution is proposing a novel way to generate synthetic clinical notes using a pre-trained large language model (e.g., GPT-3.5, GPT-4), where we create note-generating prompts by extracting key phrases from the real notes. A comparative experiment shows that the synthetic clinical notes closely follows the performance of the real notes even after using fewer number of key-phrases.

(3) *Utility evaluation of synthetic clinical notes* As there is no consensus on the utility metric of synthetic clinical notes, our third contribution is an empirical investigation of whether synthetic clinical notes can achieve comparable performance to real clinical notes. In this regard, we used the much-explored ICD code assignment (i.e., classification) problem[31] as the target task. The ICD-9 code refers to the Ninth Revision of the International Classification of Diseases (ICD), a system used to code various diseases, conditions, and medical procedures. The ICD-9 code set was widely utilized for describing patient diagnoses, symptoms, and medical billing. Manual ICD-9 coding requires coders to have a specialized understanding of medicine, coding regulations, and medical terminologies[32]. We measured the utility of the synthetic notes in terms of the performance of the underlying classifier against real notes when the classifier is trained using synthetic notes. Our experimentation showed that synthetic clinical notes attained a performance level close to real notes.

(4) *Privacy evaluation of synthetic clinical notes* Lastly, we demonstrated how a membership inference attack can be mounted when the underlying model is trained with synthetic textual data. Our proposed key-phrase-based synthetic note-generation technique follows a similar utility-privacy trade-off pattern observed in the real notes. Our literature review found no work that performed a membership inference attack on a machine-learning model trained with synthetic clinical notes. The novelty here is how the membership inference attack is mounted.

In summary, we not only demonstrated the failure of de-identification to protect against MIA attacks but also made significant contributions to enhancing the methodology for generating synthetic clinical notes and evaluating their utility and privacy aspects. The rest of this article is organized as follows. "Results" section contains two sections: "The membership inference attack on the real notes", and "Performance of the synthetic notes". The first subsection presents the results from mounting the membership inference attacks on the ICD-9 classifier instances. The impact of reducing the training-set size of the victim classifier on the success of the membership inference attack was investigated. We also investigated the impact of reducing the note length (i.e., less disclosed data) on the membership inference attack. The effect of these reductions on the ICD-9 classifiers' performance is also shown, revealing the reduction mechanism's utility-privacy trade-off. The second subsection presents the ICD-9 classification performance (on real test data) when instances of the classifier were trained and validated with the synthetic clinical notes. Similar to the previous subsection, we also present the utility and the privacy aspects and compare them with the real notes. The "Discussion" section explores the implications of this study and outlines potential future directions. Finally, in the "Methods" section, we discussed our proposed techniques and the implementation details.

## Results

For all purposes of this study, we utilized the discharge summaries from the MIMIC-III dataset, which was de-identified. A discharge summary consolidates all pertinent information from a patient's hospitalization into one comprehensive unstructured text document. Human coders assigned ICD-9 codes to each discharge summary. For synthetic note generation, we utilized the HIPAA-compliant Azure OpenAI platform provided by UTHealth in order to adhere to MIMIC-III's data usage agreement requirement. Following the benchmark work for the utility aspect (ref. Table 2 in Vu et al.[33]), we limit ourselves to the discharge summary clinical notes and focus on the 50 most frequent ICD-9 codes as the classification targets. To assess the performance of the synthetic notes, micro- and macro-averaged values for AUC, precision, recall, and F1-score of the ICD-9 classifier were reported. Each pair of (clinical note, ICD-code) is treated as a distinct prediction to calculate the micro-averaged values. On the other hand, macro-averaged values are determined by computing metrics for each label and then averaging them. Classification precisions for top-k ICD-9 codes were also reported (k=1, 5, 8, 10, and 15). We mounted the membership inference attack as a privacy measurement technique (using the TensorFlow Privacy library[34] released by Google; the details are discussed in the "Methods" section). We used three attack models (membership inference classifiers) on real data: the K-nearest neighbor model, the multi-layer perceptron model, and the random forest model. Similar to[11], the random forest attack model achieved the highest attacker advantage. We used the results from the random forest attacker model for subsequent reporting. To assess the performance of the membership inference attack, we report the attacker's advantage $(true_positive_rate(TPR) - false_positive_rate(FPR))$ and the AUC.

### Membership inference attack on the real notes and the trade-off

Our crucial finding is that the MIA attack was significantly successful despite the ICD-9 classifier being trained with de-identified real data. For the full training dataset with a maximum length of 4000 tokens for each note, the attacker had an advantage of 0.47 with an AUC of 0.79 (Fig. 1). Table 1 presents the performance of the ICD-9 code classifier and the membership attacker's advantage on different training set sizes and note lengths. The ICD-9 classification score (F1) and the attacker's advantage diminish gradually when the training-set size and the length of the notes decrease. However, this reduction in the attacker's advantage came at the price of a drop in the ICD-9 classifier's performance. Table 2 shows the gradually declining precision with the reduction in the note length and the training set size.

From these two tables, we observe the usual privacy-utility trade-off at play on the ICD-9 classifier. To reiterate our finding, as MIMIC-III clinical notes were already de-identified, this result demonstrates that membership inference attacks may become successful despite the absence of primary health identifier (PHI) tokens in the clinical notes. In other words, removing PHI tokens is not sufficient to provide full privacy.

### Performance of the synthetic notes

The study[33] we used as the benchmark utilized a window of 4000 tokens (byte-pair encoded). Due to using GPT-3.5-turbo, which is restricted to 4097 tokens (including both prompt and completion), we allocated a maximum of 3000 tokens for generating the synthetic notes. We reduced the training set to half (4033 notes) to reduce the time and the generating costs. We consider the performance of the benchmark classifier on this real dataset (4033 notes in the training set, note length of a maximum of 3000 tokens) as the baseline for our experiment.

Our proposed algorithm for generating synthetic clinical notes has been described in the algorithm 1 in "Methods" section. The algorithm relies on extracting key phrases from real notes to prepare a prompt for the synthetic note generation. (Tables 3 and 4 present the performance of the ICD-9 code classifier on notes generated under three different amounts of key phrases. The synthetic notes show a small drop of 1–3 points in micro F1 and 1–4 points in macro F1 score. Similarly, the attacker's advantage dropped by 2–5 points. In a similar study unrelated to the healthcare sector, GPT-2-large generated synthetic product reviews (of merely 100 tokens) dropped 16 points in accuracy when making predictions for a 45-class attribute, A2[35]. Our proposed approach of generating synthetic notes produces better notes with superior accuracy (a drop of $\approx$ 4–6 points) than the previous approaches. However, this synthetic note-generation approach has not yet solved the privacy dilemma. It demonstrates a clear privacy-utility trade-off where more key phrases produce better utility at the expense of less privacy.

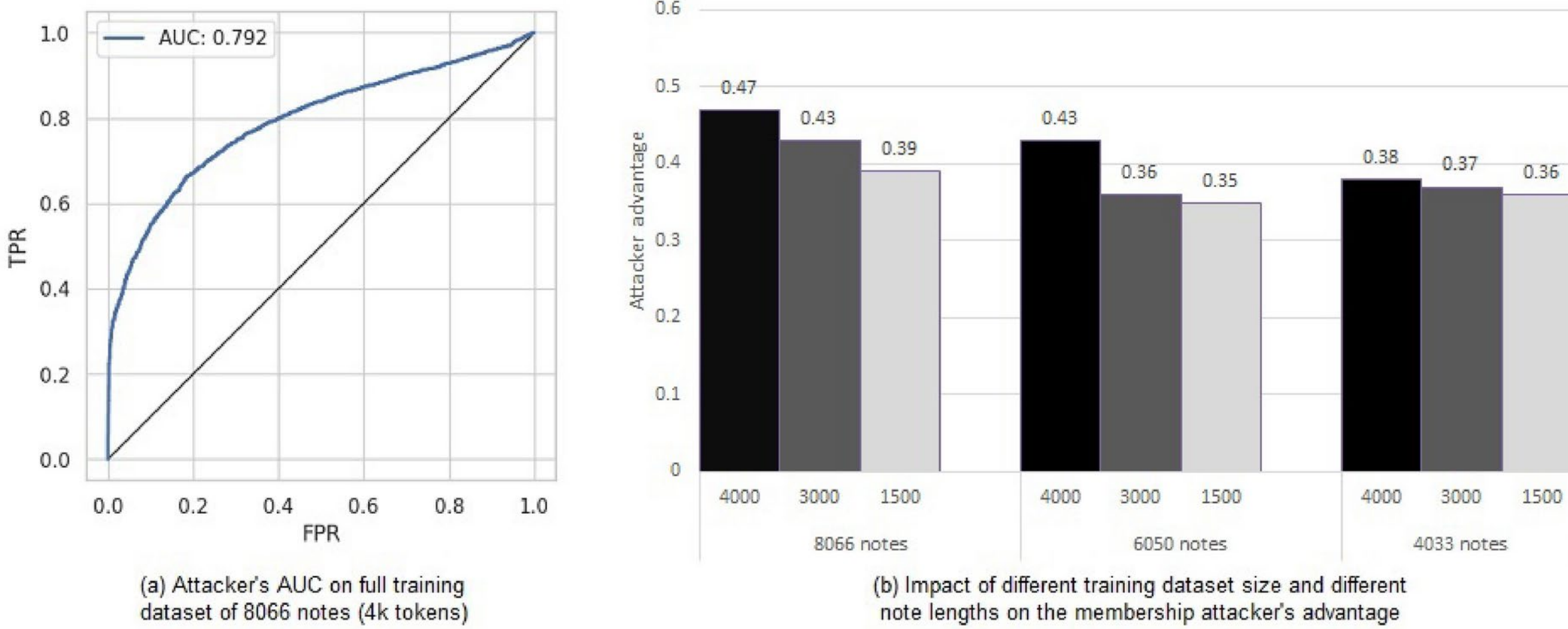

**Fig. 1**. Left: attacker's AUC on the full training set and full length notes; right: Attacker's advantage on varying training set size and note lengths.

| Notes in training set | Tokens in training set (max) | ICD-9 code classification | | | | | MIA attack | |
|---|---|---|---|---|---|---|---|---|
| | | Stat. type | AUC | Precision | Recall | F1 | Attacker advantage | AUC |
| 8066 | 4000 | Micro | 0.94 | 0.75 | 0.69 | 0.71 | 0.47 | 0.79 |
| | | Macro | 0.92 | 0.69 | 0.63 | 0.66 | | |
| | 3000 | Micro | 0.94 | 0.74 | 0.69 | 0.71 | 0.43 | 0.78 |
| | | Macro | 0.92 | 0.69 | 0.63 | 0.66 | | |
| | 1500 | Micro | 0.92 | 0.74 | 0.59 | 0.65 | 0.39 | 0.76 |
| | | Macro | 0.89 | 0.66 | 0.52 | 0.58 | | |
| 6050 | 4000 | Micro | 0.94 | 0.73 | 0.68 | 0.70 | 0.43 | 0.77 |
| | | Macro | 0.92 | 0.68 | 0.62 | 0.65 | | |
| | 3000 | Micro | 0.94 | 0.73 | 0.68 | 0.70 | 0.36 | 0.74 |
| | | Macro | 0.91 | 0.68 | 0.62 | 0.65 | | |
| | 1500 | Micro | 0.92 | 0.73 | 0.57 | 0.64 | 0.35 | 0.72 |
| | | Macro | 0.89 | 0.65 | 0.51 | 0.57 | | |
| 4033 | 4000 | Micro | 0.94 | 0.73 | 0.68 | 0.70 | 0.38 | 0.75 |
| | | Macro | 0.92 | 0.68 | 0.62 | 0.65 | | |
| | 3000 | Micro | 0.93 | 0.74 | 0.65 | 0.69 | 0.37 | 0.75 |
| | | Macro | 0.91 | 0.67 | 0.60 | 0.63 | | |
| | 1500 | Micro | 0.91 | 0.67 | 0.62 | 0.64 | 0.36 | 0.74 |
| | | Macro | 0.88 | 0.61 | 0.55 | 0.58 | | |

**Table 1**. The performance of the ICD-9 code classifier (frequent-50) and the membership attacker's advantage under different training (member) set sizes and note lengths.

| Notes in training set | Tokens in training set (max) | P@1 | P@5 | P@8 | P@10 | P@15 |
|---|---|---|---|---|---|---|
| 8066 | 4000 | 0.88 | 0.67 | 0.54 | 0.48 | 0.36 |
| | 3000 | 0.88 | 0.67 | 0.54 | 0.47 | 0.35 |
| | 1500 | 0.85 | 0.63 | 0.51 | 0.45 | 0.34 |
| 6050 | 4000 | 0.86 | 0.66 | 0.54 | 0.48 | 0.36 |
| | 3000 | 0.88 | 0.66 | 0.54 | 0.47 | 0.35 |
| | 1500 | 0.85 | 0.63 | 0.51 | 0.45 | 0.34 |
| 4033 | 4000 | 0.88 | 0.66 | 0.54 | 0.47 | 0.35 |
| | 3000 | 0.86 | 0.66 | 0.53 | 0.47 | 0.35 |
| | 1500 | 0.84 | 0.61 | 0.50 | 0.44 | 0.34 |

**Table 2**. Impact of varying the training dataset size and the note size on the precision of top-k ICD-9 code classification (dataset: real notes; k = 1, 5, 8, 10, 15).

## Discussion

De-identification is often considered sufficient to ensure privacy. For example, the office of the *Information and Privacy Commissioner* of Ontario stated on their website the following[36].

> "De-identification protects the privacy of individuals because once de-identified, a data set is considered to no longer contain personal information. If a data set does not contain personal information, its use or disclosure cannot violate the privacy of individuals. Accordingly, the privacy protection provisions of the Freedom of Information and Protection of Privacy Act and Municipal Freedom of Information and Protection of Privacy Act would not apply to de-identified information."

We trained a machine learning classifier with de-identified clinical notes. We mounted a membership inference attack on the classifier and found the attack quite successful. This finding has far-reaching privacy implications for membership-sensitive models that use clinical notes in their training. The concern is that if an adversary obtains a target person's clinical note, they could use a membership inference attack to determine with reasonable probability whether the person's note was in the training dataset of the sensitive machine learning model. For example, similar to the re-identification scenarios mentioned by the European Medicines Agency[37], a person can be of significant public interest, leading to focused attention from the press or other entities. The adversary may plot a devious plan to get the clinical note and subsequently mount MIA attack. Or perhaps a random event occurs where an individual, familiar with a target participant came across the note and then tries to know the membership status. As a proactive precautionary step, our suggestion would be to refrain from publishing membership-sensitive models with full access to prediction probabilities until a better defense against this attack is developed.Our another main contribution is that we proposed a new approach to generate synthetic clinical notes and investigated whether these synthetic notes nullifies the membership inference attack. The pre-trained large language models, such as GPT-3.5, can generate synthetic clinical notes corresponding to the real notes. We have demonstrated that simply extracting key phrases from de-identified notes and instructing GPT to produce notes with those key phrases provide comparable utility like real notes but does not solve the membership inference attack. An important research topic in the future could be to investigate which features of the notes help distinguish between members and non-members in membership inference attacks. One interesting takeaway from our experiments is that reducing the number of key phrases to one-fourth (1000–250) does not drastically destroy the utility. This is a promising finding, as manipulating a small number of key phrases could solve this privacy dilemma, providing a better utility-privacy trade-off.

| | | ICD-9 code classification | | | | | MIA attack | |
|---|---|---|---|---|---|---|---|---|
| Model | Number of key-phrases extracted by KP-miner and YAKE for the prompt | Stat. type | AUC | Precision | Recall | F1 | Attacker advantage | AUC |
| Benchmark | – | Micro | 0.93 | 0.74 | 0.65 | 0.69 | 0.37 | 0.75 |
| | | Macro | 0.91 | 0.67 | 0.60 | 0.63 | | |
| GPT 3.5 turbo 0301 | 1000 | Micro | 0.93 | 0.73 | 0.62 | 0.67 | 0.35 | 0.73 |
| | | Macro | 0.90 | 0.64 | 0.55 | 0.59 | | |
| | 500 | Micro | 0.93 | 0.72 | 0.62 | 0.67 | 0.34 | 0.73 |
| | | Macro | 0.90 | 0.66 | 0.56 | 0.60 | | |
| | 250 | Micro | 0.93 | 0.75 | 0.59 | 0.66 | 0.32 | 0.71 |
| | | Macro | 0.90 | 0.68 | 0.51 | 0.59 | | |
| GPT 3.5 turbo 0613 | 1000 | Micro | 0.93 | 0.72 | 0.64 | 0.68 | 0.35 | 0.74 |
| | | Macro | 0.90 | 0.67 | 0.57 | 0.62 | | |
| | 500 | Micro | 0.93 | 0.75 | 0.61 | 0.67 | 0.34 | 0.71 |
| | | Macro | 0.90 | 0.67 | 0.54 | 0.59 | | |
| | 250 | Micro | 0.93 | 0.71 | 0.62 | 0.66 | 0.34 | 0.72 |
| | | Macro | 0.90 | 0.69 | 0.56 | 0.62 | | |

**Table 3**. The performance of the synthetic notes on ICD-9 code classification and the membership inference attack.

| Model | Number of key-phrases extracted by KP-miner and YAKE for the GPT prompt | P@1 | P@5 | P@8 | P@10 | P@15 |
|---|---|---|---|---|---|---|
| Benchmark | – | 0.86 | 0.66 | 0.53 | 0.47 | 0.35 |
| GPT 3.5 turbo 0301 | 1000 | 0.85 | 0.64 | 0.53 | 0.46 | 0.35 |
| | 500 | 0.85 | 0.64 | 0.52 | 0.46 | 0.34 |
| | 250 | 0.85 | 0.64 | 0.52 | 0.46 | 0.35 |
| GPT 3.5 turbo 0613 | 1000 | 0.84 | 0.64 | 0.53 | 0.46 | 0.35 |
| | 500 | 0.85 | 0.65 | 0.53 | 0.46 | 0.35 |
| | 250 | 0.84 | 0.66 | 0.55 | 0.49 | 0.37 |

**Table 4**. Impact of varying the amount of key phrases in the prompt on the precision of top-k ICD-9 code classification (test dataset: real notes; k = 1, 5, 8, 10, 15).

The current landscape of pre-trained large language models is rapidly changing. The limitation in context windows that previously prohibited generating larger notes (e.g., 1024 tokens for GPT-2) is being overcome rapidly (e.g., 32k context length in GPT-4-32k)[38]. As researchers and engineers continue to push the boundaries of language models, new architectures and training techniques are emerging to address the limitations of previous models. As a result, they will generate more accurate and contextually appropriate responses, making them increasingly valuable for real-world use cases. As the field continues to evolve, researchers should try to unlock the full potential of large language models, leading to groundbreaking and transformative applications.

In conclusion, our research demonstrates the vulnerability of a machine learning model even when trained with de-identified data. We also provided a new algorithm for synthetic clinical note generation that performs comparably to real notes in a clinical domain task. The relatively small number of key phrases required to generate the synthetic notes might hold the key to the solution of the membership inference attack. Future research should continue to explore this promising approach.

## Methods

The experimentation conducted in this paper relies on the methods presented in this section, which is divided into three subsections. The first subsection describes our proposed algorithm for generating synthetic clinical notes. The next subsection briefly describes a benchmark ICD-code classifier from the literature, which we used to measure the utility of the synthetic notes. The final subsection describes (i) the membership inference attack applied to the ICD-code classifier when trained with real data and (ii) our proposed attack strategy when the ICD-code classifier is trained with synthetic data.

### Synthetic clinical note generation

Our proposed algorithm for synthetic clinical note generation consists of three main steps: (i) Keyword/key-phrase extraction, (ii) prompt creation, and (iii) note generation. For large language models, the quality of the generated text is primarily determined by the quality of the provided prompt. Our algorithm utilized key-phrase extraction to create the prompt for synthetic clinical note generation. Keyword/key-phrase extraction involves automatically extracting single or multiple-token phrases from a textual document that most effectively captures all essential aspects of its content. This process can be viewed as the automated generation of a concise document summary. Key-phrase extraction often serves as the foundational technology for document indexing, clustering, classification, and summarization[39]. The reasoning behind utilizing key phrases in the prompt for synthetic clinical note generation is that, given key-phrases, synthetic note generation can be considered as the reverse task of key-phrase extraction. We hypothesized that a prompt containing the key-phrases in a certain orderly fashion (e.g., sorted by location of their appearance in the original note) would help the GPT to come up with a synthetic note that is similar to the original note in clinical aspects but different in the linguistic style (as many parts of the original note are not present in the key-phrases). We expected that since GPT will generate the synthetic note with somewhat different sentences than the original notes while maintaining the critical information pieces (i.e., key-phrases) in a similar order, it will provide utility similar to the original note while membership-privacy may be gained through the difference in the linguistic style of the sentences. It seemed a reasonable assumption considering a literary person may infer whether a text snippet is written by Shakespeare or not.

```
1:  Extract key-phrases from the real notes. (e.g., KPminer, YAKE!)
2:  [Optional] De-duplicate the key-phrases (e.g., using the FuzzyWuzzy library)
3:  if multiple key-phrase extraction technique is used (e.g., KP-miner and YAKE simultaneously) then
4:          Sort them based on their appearance in the real note, merge the key-phrases from all the extractions (process
            overlapping appearances) and use them in the generation prompt.
5:  else[One key-phrases extraction technique is used]
6:          Sort them based on their appearance in the real note and use them in the generation prompt.
7:  end if
8:  if fine-tuning is required then
9:      Fine-tune the pre-trained large language model
10: else:
11:     Use the pre-trained large language model without fine-tuning
12: end if
```

**Algorithm 1**. Synthetic clinical note generation using a large language model (e.g., GPT-3.5, GPT-4)

*Subtask: key-phrase extraction* We used two key-phrase extraction algorithms, KP-miner and YAKE, to create prompts for generating synthetic notes. Both KP-Miner and YAKE are unsupervised keyword extraction algorithms that use certain statistical features of a document. KP-Miner leverages both the frequency and positional information of candidate keywords, incorporating a weighting mechanism for multi-token keywords. YAKE uses various statistics of the candidate keywords related to the position, frequency, context, etc., to create a score to rank the candidate keywords. In a survey[40], the statistical technique KP-miner has been found to be superior over nine other key-phrase extraction techniques (on tasks: Semeval, NUS, Krapivin). Another study[39] found the best performance in indexing news articles using KP-miner and YAKE combinedly. We used open-source libraries for both KP-miner[41] and YAKE[42].

*Subtask: key-phrase de-duplication* For the de-duplication of key-phrases extracted by KP-miner, we utilized an open-source function from the fuzzywuzzy package[43] that uses Levenshtein distance in the process for measuring similar key-phrases.

*Subtask: using an LLM for note generation* The purpose of fine-tuning an LLM for clinical note generation is to adapt the model to specialized medical terminology, patient information context, structure, and domain knowledge. However, fine-tuning is not always necessary if the model's pre-trained knowledge closely aligns with the desired output. In cases where only minimal structural adaptation is needed, prompt engineering may be sufficient to guide the LLM effectively. Additionally, fine-tuning may not be feasible if the LLM is hosted by a third party and privacy regulations prohibit sending private notes.

*Implementation detail*
We first preprocess the notes for keyword extraction according to[33]. It removed numeric data from the notes and made all words lowercase. For key-phrases extraction using the KP-miner, we set its least allowable seen frequency (lasf) parameter to 1. For the YAKE algorithm, we kept the maximum n-gram size to default 3 and the de-duplication function to the default 'sequencematcher' function with a threshold of 0.70. After getting the keywords, we sort them based on the place of appearance in the note. We use this sorted sequence of phrases to merge and remove duplicate phrases. Finally, we use this list of phrases in the prompt for note generation.

We used the same set of key-phrases for gpt-3.5-turbo-0301 and gpt-3.5-turbo-0613. The difference in the prompts is: in version-0301, the instruction that precedes the list of key phrases is "*Write a description of a patient using 2250 words containing the following phrases sequentially:[... comma-separated list of phrases...]*". For gpt-3.5-turbo-0613, the preceding instruction was "*Write a description of a patient using 2250 words. The description should contain the following phrases sequentially: [... comma-separated list of phrases...]*".

## Utility: ICD coding from clinical text

To compare the synthetic notes' utility aspect, we used the ICD-9 code classifier proposed in[33]. This model employs a bidirectional Long-Short Term Memory (BiLSTM) encoder to learn label-specific vectors representing essential clinical text fragments related to specific ICD labels. They proposed a novel label attention mechanism. The final layer involves an array of 50 binary classifiers, one for each of the 50 most occurring ICD-9 codes. Figure 2 presents the architecture of their model (they name it LAAT, short for Label Attention).

At the time of its publication, this model was state-of-the-art (SOTA), and currently, it is slightly behind the new SOTA (Precision@15: 59.1 vs. 61.5 on the full test set; micro-F1: 57.5 vs. 59.9)[31]. We adopt this model for the benchmark due to its open-source nature and simple architecture.

## Privacy: membership inference attack

*Threat model*
Our threat model is similar to the model described by Salem et al. (ref. section III, page 4[14]). The attacker is a supervised binary classifier, distinguishing between members and non-members. The adversary has access to the victim's original record and the knowledge of PHI fields (It is a reasonable assumption as HIPAA lists 18 categories of PHI attributes and notes are usually de-identified on those attributes). The adversary may or may not know the particular PHI de-identification algorithm used by the victim model, but knowing about the PHI token fields makes it possible to de-identify the note. To facilitate the training of this attack model, the adversary needs to obtain a labelled training dataset (clinical notes with ground truth membership information). The adversary accomplishes this by training a shadow model using a shadow dataset sourced from the identical underlying distribution to the target model's training data. This approach aims to emulate the behaviour of the target model, relying on the shadow model to acquire the ground truth membership required for training the attack model. The shadow model is assumed to operate under the same algorithm and architecture as the target model. When using synthetic notes, it is assumed that the attacker also knows the synthetic note generation approach used by the victim classifier.

Figure 3 shows the conceptual architecture of the attack. It involves two classifiers: the victim classifier (in this experiment, the ICD-9 code classifier) and the membership-inference classifier (in this experiment, the MIA classifier from the TensorFlow-privacy library). The assumption is that the victim classifier will behave differently to training data than non-training data when it gets them as input. For example, it may predict the class label more confidently (i.e., higher probability). The attacker's classifier is trained to capture this behaviour and distinguish between members and non-members, given the class prediction vector. Figure 3b shows how the attacker's inference classifier is trained when the victim classifier is trained with real data. At first, the attacker trains a shadow model of the victim classifier with shadow data. After training, the attacker derives class prediction vectors for the shadow training dataset ($D_{Shadow}^{Train}$, label: member) and the shadow non-training dataset ($D_{Shadow}^{Out}$, label: non-member). The attacker then trains a membership inference classifier on these labelled class prediction vectors. After the training is completed, the attacker uses the target's prediction vector from the victim classifier as input to infer the target's membership.

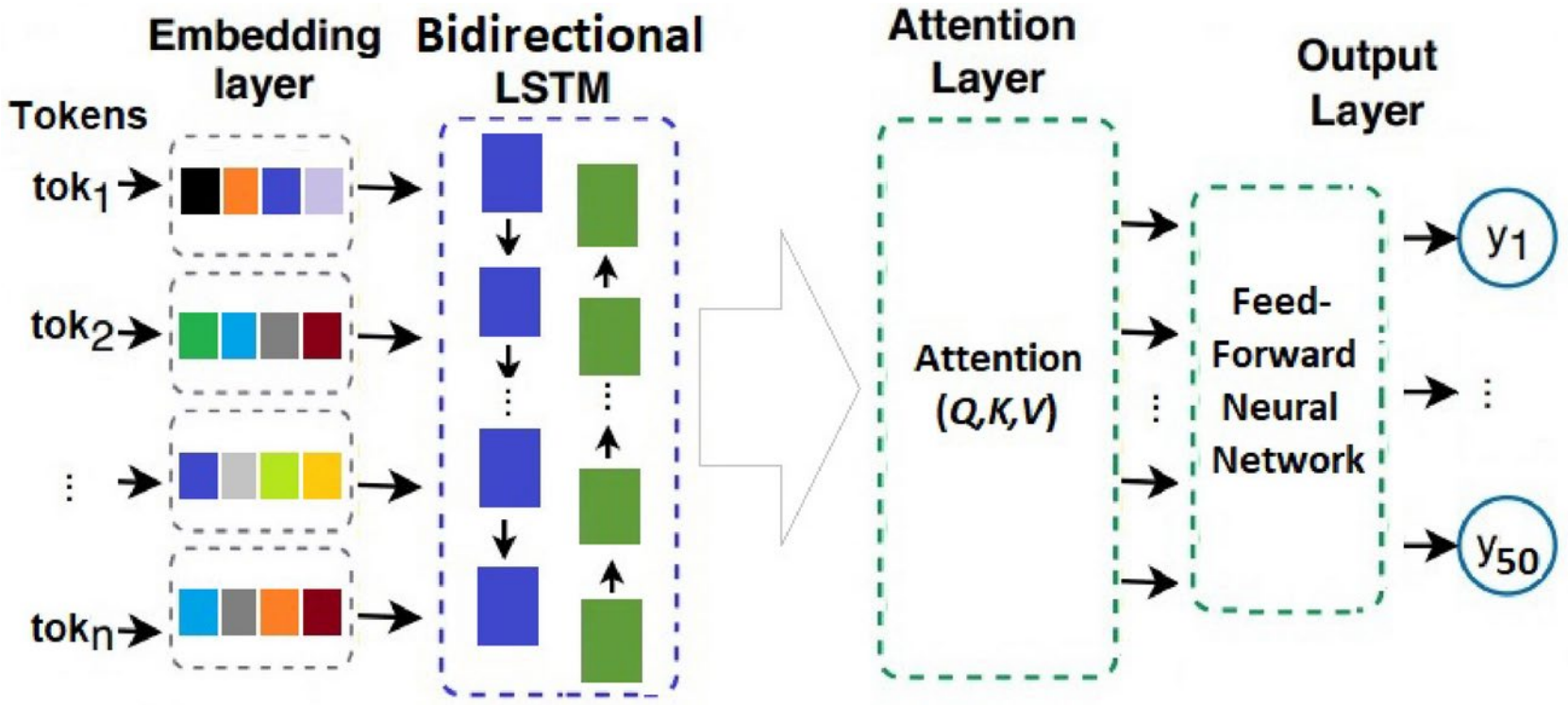

**Fig. 2**. Architecture of the ICD-9 classifier (LAAT model, figure adapted from[33]).

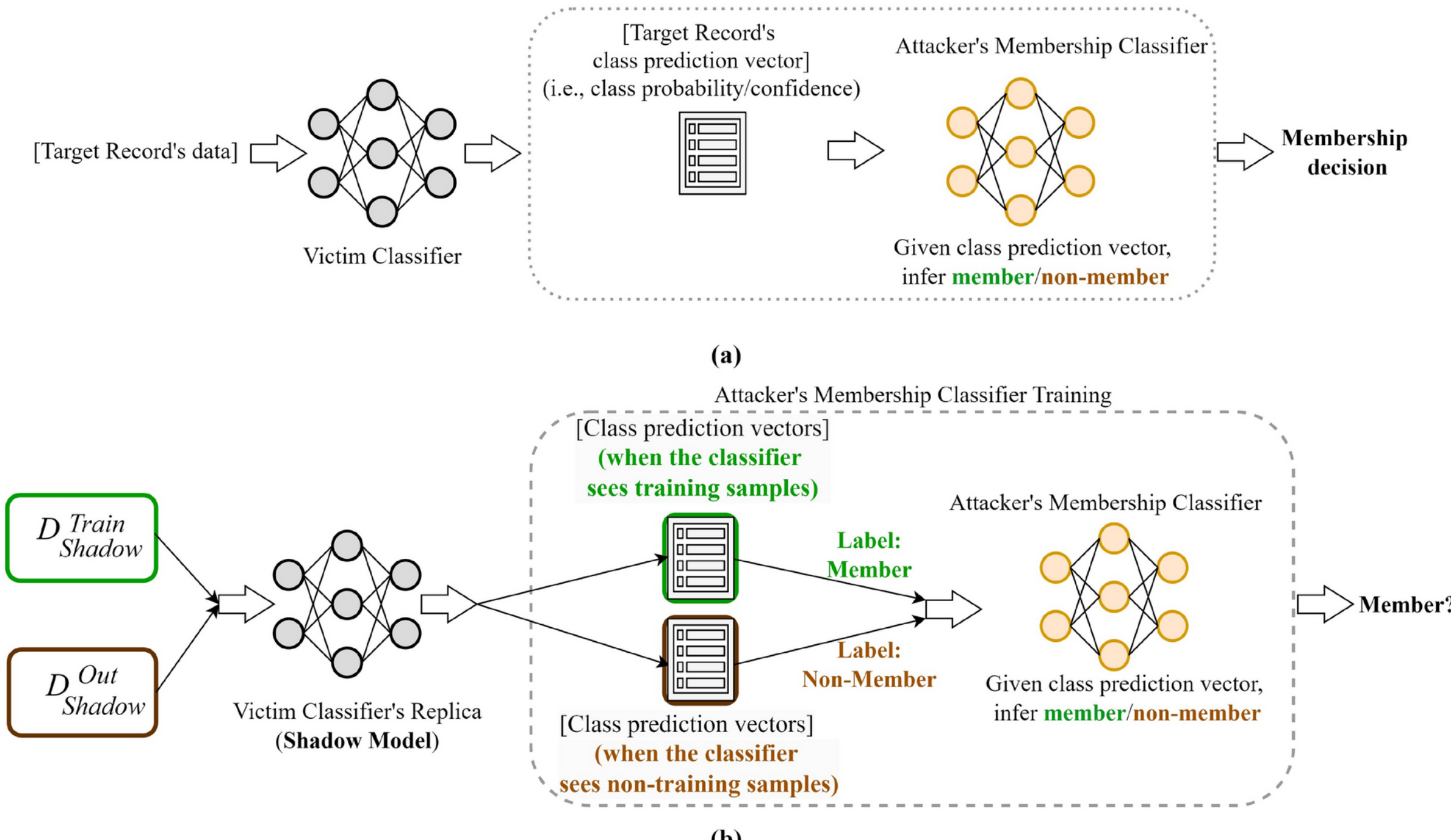

**Fig. 3**. Simplified pictorial representation of the MIA attack: (**a**) typical attack scenario: the attacker uses the prediction probabilities from the victim classifier to infer the target's membership, (**b**) steps for training the attacker's inference classifier when the victim model is trained with real data: the attacker builds a shadow model and gathers the probabilities for the member (i.e., training set) and non-member records. The attacker then uses this probabilities to train a membership classifier.

Figure 4 shows our proposed adaptation to how the attacker's inference classifier is trained when the victim classifier is trained with synthetic data. In this scenario, there is a synthetic data generator that takes real data as input and produces synthetic data as output which is then used by the victim classifier (e.g., the ICD-9 code classifier) to train itself. The attacker assumes that the victim's classifier, trained on synthetic data, may behave differently when it sees the real data that was used to train the synthetic data generator. Similar to the earlier scenario, the attacker gets two shadow datasets ($D^{Train}_{Shadow}$ marked in a green box and $D^{Out}_{Shadow}$ marked in a brown box). The $D^{Train}_{Shadow}$ dataset is used to train a synthetic data generator, and a synthetic dataset sampled from this synthetic data generator is used to train the shadow victim model (i.e., the synthetic dataset is used to train the victim classifier, ICD-9 code classifier). Then, the attacker derives class prediction vectors from the victim model for the $D^{Train}_{Shadow}$ dataset (label: member) and non-training dataset $D^{Out}_{Shadow}$ (label: non-member). The attacker then trains the membership inference classifier (e.g., the MIA classifier from the TensorFlow-privacy library) on these labelled class prediction vectors and uses this trained MIA classifier to infer the target's membership.

*Experimental setup for tensorflow-privacy's MIA*

In this work, we used TensorFlow Privacy library's membership inference attack as released by Google Research[34]. It has a different approach than the original attack proposed by Shokri et al.[11] The attack does not involve training multiple shadow models as described in the original attack. Instead, it leverages the findings reported by Salem et al.[14] Salem et al.'s approach uses one shadow model[14], eliminating the need to train multiple shadow models that approximate the original model's behavior. In the TensorFlow Privacy's MIA attack, the original model being targeted acts as the shadow model, perfectly approximating its behaviour. The purpose of a shadow model is to emulate the behavior of the victim model as closely as possible so that the attacker can acquire the ground truth membership required for training the attack model (i.e., prediction vectors of the training set are 'member' and prediction vectors of the testing set are 'non-member'). Suppose the owner of a (victim) model wants to assess its MIA privacy state. To capture the worst-case leakage, the shadow model needed to behave the same as the target model. The TensorFlow-privacy library's MIA attack does this exact replication of behaviour by making disjoint subsets of the victim's training dataset and uses these disjoint subsets to act as both the shadow dataset and the target dataset (but not simultaneously, i.e., when one subset acts as the shadow dataset, another subset acts as the target dataset). When the attacker is trained, an equal number of member and non-member examples is chosen for the attacker training similar to Salem et al.[14] (if member and non-member datasets are unequal in size, the attacker's dataset size becomes the fewer of the two; uniform random sampling without replacement is used on the larger class). To get the membership scores of all samples, the experiment is done multiple times in a way that each example gets its score assigned only once. It uses StratifiedKFold (cross-validation folds = 2 by default) for this purpose. When the member and non-member datasets are unequal in size, the left-out samples of the larger dataset are evaluated using the last attacker. We believe that this approach in the TensorFlow-privacy's MIA attack captures the worst leakage possible under Salem et al.'s MIA attack. It is a way to follow Murphy's law: "Anything that can go wrong will go wrong and at the worst possible time."

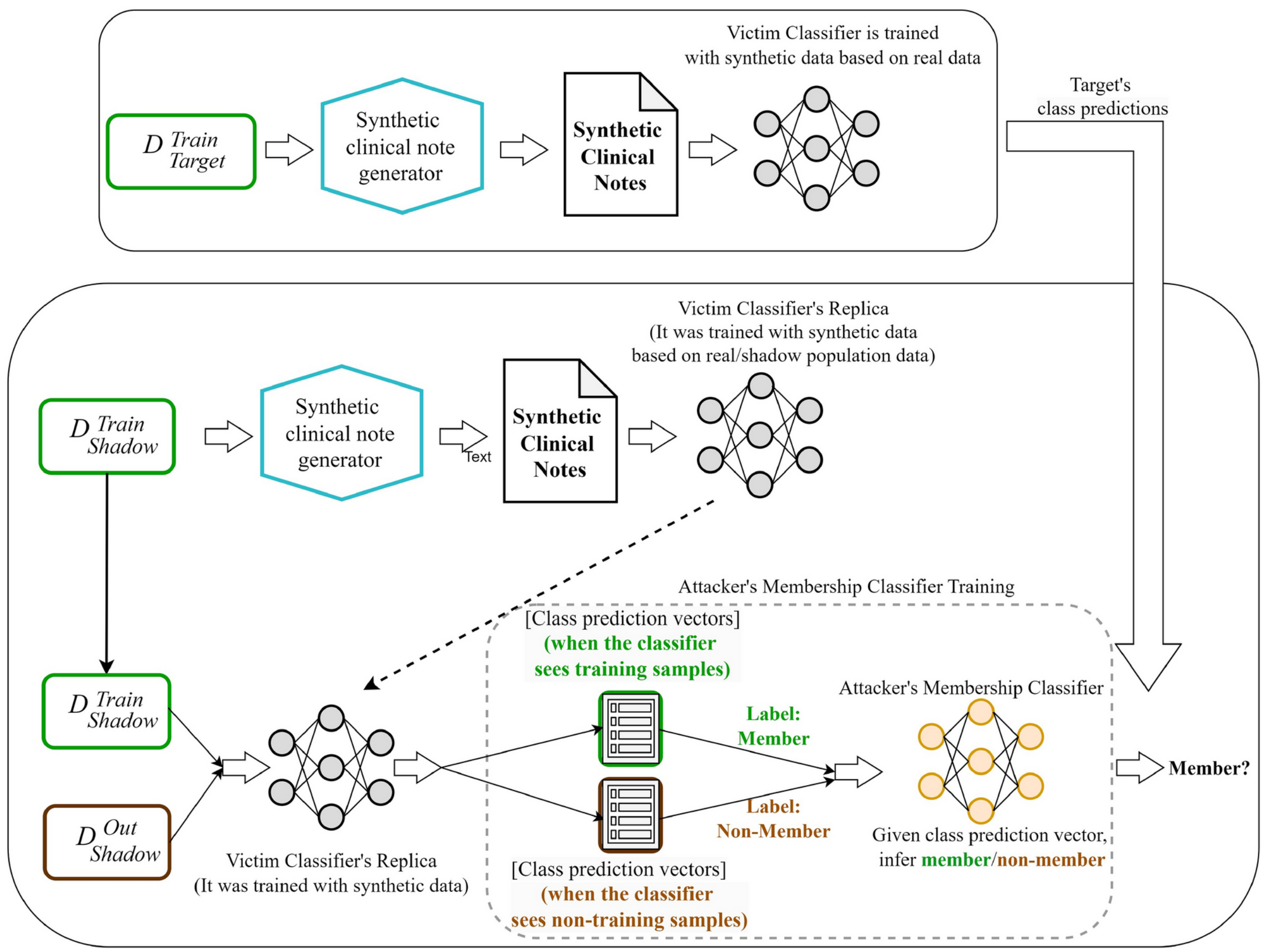

**Fig. 4.** Simplified pictorial representation of the MIA attack steps when the victim model is trained with synthetic data.

Let us elaborate on the process with an example. Suppose the owner of the victim model has 4033 clinical notes for training and 1729 clinical notes for testing. The owner wishes to measure the susceptibility of the ICD-9 classifier to an MIA attack. The owner first trains the ICD-9 classifier on 4033 notes. Then, the owner gets the class prediction vectors of training data (4033 notes, label: member) and testing data (1729 notes, label: non-member). The owner then fed these labelled class prediction vectors (size: $4033 \times 50$ and $1729 \times 50$) to the TensorFlow-privacy's MIA attack module. Since the member and non-member datasets are unequal in size, the attacker's dataset contains 1729 member and 1729 non-member samples, where the 1729 'member' samples were chosen by uniform random sampling without replacement from the 4033 samples). This resulted in 2304 left-out member samples. These 1729 member and 1729 non-member samples are used to train and test the attacker classifier via stratified 2-fold unshuffled sampling (In each fold, the attack model receive half of the 1729 member and 1729 non-member prediction vectors. The testing phase involves the other half of the 1729 member and 1729 non-member prediction vectors.). The second attack model (i.e., the attacker from the 2nd fold) is used to infer the membership of the 2304 leftout member samples, resulting in all the samples (1729 + 1729 + 2304) getting their membership assigned exactly once.(For details, check the $_run_trained_attack()$[44])

To train the attacker's inference classifier when the victim classifier (ICD-9 classifier) was trained with synthetic data, we first generated synthetic notes based on the real train and validation data. Then, we trained the ICD-9 classifier on these synthetic data. Then, we got the class prediction vectors of the real training data (label: member) and the real testing data (label: non-member). We then fed these labelled class prediction vectors to the random forest classifier for training as the attacker's inference classifier. For replication purposes, here we describe the parameters that we configured. In the AttackInputData, we set "*multilabel_data = True*" because the ICD-9 classifier works with 50 classes. In the SlicingSpec, we set "*entire_dataset = True*". Our attacks included threshold attack, random forest model, logistic regression, KNN, and multi-layer perceptron (set in *attack_types*). During "*run_attacks*", we set "*balance_attacker_training = True*".

## Data availability

MIMIC-III: is a publicly available dataset (https://doi.org/10.13026/C2XW26) but requires signing a data us age agreement (DUA) and a specific training. Available through: https://physionet.org/content/mimiciii/1.4/. Azure-OpenAI platform: The Azure-openAI platform is HIPAA compliant while the publicly available openAI platform is not. We utilized the Azure OpenAI platform provided by UTHealth as MIMIC-III requires HIPAA compliance (more information on https://physionet.org/news/post/415.

## References

1. Boag, W., Doss, D., Naumann, T. & Szolovits, P. What's in a note? Unpacking predictive value in clinical note representations. *AMIA Summits Transl. Sci. Proc.* **2018**, 26 (2018).
2. Van Aken, B. et al. Clinical outcome prediction from admission notes using self-supervised knowledge integration. *arXiv preprint*[SPACE]arXiv:2102.04110 (2021).
3. Ye, J., Yao, L., Shen, J., Janarthanam, R. & Luo, Y. Predicting mortality in critically ill patients with diabetes using machine learning and clinical notes. *BMC Med. Inform. Decis. Mak.* **20**, 1–7 (2020).
4. Payrovnaziri, S. N., Barrett, L. A., Bis, D., Bian, J. & He, Z. Enhancing prediction models for one-year mortality in patients with acute myocardial infarction and post myocardial infarction syndrome. *Stud. Health Technol. Inform.* **264**, 273 (2019).
5. Cai, X. et al. Real-time prediction of mortality, readmission, and length of stay using electronic health record data. *J. Am. Med. Inform. Assoc.* **23**, 553–561 (2016).
6. El Emam, K., Mosquera, L. & Bass, J. Evaluating identity disclosure risk in fully synthetic health data: model development and validation. *J. Med. Internet Res.* **22**, e23139 (2020).
7. Ahmed, T., Aziz, M. M. A. & Mohammed, N. De-identification of electronic health record using neural network. *Sci. Rep.* **10**, 18600 (2020).
8. Urbain, J. et al. Natural language processing for enterprise-scale de-identification of protected health information in clinical notes. In *AMIA Annual Symposium Proceedings*, Vol. 2022, 92 (American Medical Informatics Association, 2022).
9. Yang, X. et al. A study of deep learning methods for de-identification of clinical notes in cross-institute settings. *BMC Med. Inform. Decis. Mak.* **19**, 1–9 (2019).
10. Scaiano, M. et al. A unified framework for evaluating the risk of re-identification of text de-identification tools. *J. Biomed. Inform.* **63**, 174–183 (2016).
11. Shokri, R., Stronati, M., Song, C. & Shmatikov, V. Membership inference attacks against machine learning models. In *2017 IEEE symposium on security and privacy (SP)*, 3–18 (IEEE, 2017).
12. Tabassi, E., Burns, K. J., Hadjimichael, M., Molina-Markham, A. D. & Sexton, J. T. A taxonomy and terminology of adversarial machine learning. *NIST IR* **2019**, 1–29 (2019).
13. Choquette-Choo, C. A., Tramer, F., Carlini, N. & Papernot, N. Label-only membership inference attacks. In *International conference on machine learning*, 1964–1974 (PMLR, 2021).
14. Salem, A. et al. Ml-leaks: Model and data independent membership inference attacks and defenses on machine learning models. *arXiv preprint*[SPACE]arXiv:1806.01246 (2018).
15. Carlini, N. et al. Membership inference attacks from first principles. In *2022 IEEE Symposium on Security and Privacy (SP)*, 1897–1914 (IEEE, 2022).
16. TensorFlow.org. Assess privacy risks with the TensorFlow Privacy Report. Available at: https://www.tensorflow.org/responsible_ai/privacy/tutorials/privacy_report (2022). Accessed 7 January 2024.
17. Murakonda, S. K. & Shokri, R. Ml privacy meter: Aiding regulatory compliance by quantifying the privacy risks of machine learning. *arXiv preprint*[SPACE]arXiv:2007.09339 (2020).
18. Hu, H. et al. Membership inference attacks on machine learning: A survey. *ACM Comput. Surv. (CSUR)* **54**, 1–37 (2022).
19. El Emam, K. & Hoptroff, R. The synthetic data paradigm for using and sharing data. *Cut. Exec. Update* **19**, 1–12 (2019).
20. Vakili, T. & Dalianis, H. Using membership inference attacks to evaluate privacy-preserving language modeling fails for pseudonymizing data. In *Proceedings of the 24th Nordic Conference on Computational Linguistics (NoDaLiDa)*, 318–323 (2023).
21. Jagannatha, A., Rawat, B. P. S. & Yu, H. Membership inference attack susceptibility of clinical language models. *arXiv preprint*[SPACE]arXiv:2104.08305 (2021).
22. Moramarco, F. et al. Human evaluation and correlation with automatic metrics in consultation note generation. *arXiv preprint*[SPACE]arXiv:2204.00447 (2022).
23. Faequa, T. *Privacy-Preserving Generation of Textual Healthcare Data* (Master's dissertation, The University of Regina, Canada, 2021).
24. Al Aziz, M. M. et al. Differentially private medical texts generation using generative neural networks. *ACM Trans. Comput. Healthc. (HEALTH)* **3**, 1–27 (2021).
25. Libbi, C. A., Trienes, J., Trieschnigg, D. & Seifert, C. Generating synthetic training data for supervised de-identification of electronic health records. *Future Internet* **13**, 136 (2021).
26. Samuel, J., Palle, R. & Soares, E. C. Textual data distributions: Kullback leibler textual distributions contrasts on gpt-2 generated texts, with supervised, unsupervised learning on vaccine & market topics & sentiment. *arXiv preprint*[SPACE]arXiv:2107.02025 (2021).
27. Zhang, T., Kishore, V., Wu, F., Weinberger, K. Q. & Artzi, Y. Bertscore: Evaluating text generation with bert. *arXiv preprint*[SPACE]arXiv:1904.09675 (2019).
28. Callison-Burch, C., Osborne, M. & Koehn, P. Re-evaluating the role of bleu in machine translation research. In *11th conference of the european chapter of the association for computational linguistics*, 249–256 (2006).
29. Li, J. et al. Are synthetic clinical notes useful for real natural language processing tasks: A case study on clinical entity recognition. *J. Am. Med. Inform. Assoc.* **28**, 2193–2201 (2021).
30. OpenAI. Gpt-4. (2023). Available at: https://openai.com/research/gpt-4 Accessed 7 January 2024.
31. Meta AI Research. Leaderboard: Medical Code Prediction on MIMIC-III. Available at: https://paperswithcode.com/sota/medical-code-prediction-on-mimic-iii. (2022). Accessed 7 January 2024.
32. Zeng, M. et al. Automatic ICD-9 coding via deep transfer learning. *Neurocomputing* **324**, 43–50 (2019).
33. Vu, T., Nguyen, D. Q. & Nguyen, A. A label attention model for ICD coding from clinical text. In *Proceedings of the Twenty-Ninth International Joint Conference on Artificial Intelligence (IJCAI-20)*, 3335–3341 (2020).
34. Google LLC. TensorFlow Privacy. Library for training machine learning models with privacy for training data. Version 0.8.8. Available at: https://github.com/tensorflow/privacy (2023). Accessed 7 January 2024.
35. Yue, X. et al. Synthetic text generation with differential privacy: A simple and practical recipe. *arXiv preprint*[SPACE]arXiv:2210.14348 (2022).

36. Information and Privacy Commissioner of Ontario. De-identification. Available at: https://www.ipc.on.ca/privacy-organizations/de-identification/ (2016). Accessed 7 January 2024.
37. European Medicines Agency, GT. External guidance on the implementation of the European medicines agency policy on the publication of clinical data for medicinal products for human use (2018).
38. Eleti, A., Harris, J. & Kilpatrick, L. Function calling and other api updates. Available at: https://openai.com/blog/function-calling-and-other-api-updates (2023). Accessed 7 January 2024.
39. Piskorski, J., Stefanovitch, N., Jacquet, G. & Podavini, A. Exploring linguistically-lightweight keyword extraction techniques for indexing news articles in a multilingual set-up. In *Proceedings of the EACL Hackashop on News Media Content Analysis and Automated Report Generation*, 35–44 (2021).
40. Papagiannopoulou, E. & Tsoumakas, G. A review of keyphrase extraction. *Wiley Interdiscip. Rev.: Data Min. Knowl. Discov.* **10**, e1339 (2020).
41. Boudin, F. Pke: an open source python-based keyphrase extraction toolkit. In *Proceedings of COLING 2016, the 26th international conference on computational linguistics: System demonstrations*, 69–73 (2016).
42. Campos, R. et al. Yake! keyword extraction from single documents using multiple local features. *Inf. Sci.* **509**, 257–289 (2020).
43. Cohen, A. Fuzzywuzzy: Fuzzy string matching in python. Available at: https://pypi.org/project/fuzzywuzzy/ (2020). Accessed 7 January 2024.
44. TensorFlow.org. Membership inference attack. Available at: https://github.com/tensorflow/privacy/blob/master/tensorflow_privacy/privacy/privacy_tests/membership_inference_attack/membership_inference_attack.py (2020). Accessed 7 January 2024.

## Acknowledgements

A.R.S. is a Gordon P. Osler scholar and was partially supported by UMGF fellowship. N.M. was supported by the NSERC Discovery Grants (RGPIN-04127-2022) and the NSERC Alliance Grants (ALLRP 592951 - 24). X.J. is CPRIT Scholar in Cancer Research (RR180012), and he was supported in part by Christopher Sarofim Family Professorship, UT Stars award, UTHealth startup, the National Institute of Health (NIH) under award number R01AG066749, R01AG066749-03S1, R01LM013712, R01LM014520, R01AG082721, R01AG084637-01A1, R01AG066749, U01AG079847, U24LM013755, U01CA274576, U54HG012510, 1OT2OD032581-02-211 and the National Science Foundation (NSF) #2124789.

## Author contributions

All authors participated in the design of the methods. A.R.S. implemented the methods in python and conducted the experiments. All authors wrote, reviewed and revised the paper. X.J. and N.M. supervised the research.

## Declarations

## Competing interests

The authors declare no competing/conflict of interests.

## Additional information

**Correspondence** and requests for materials should be addressed to A.R.S.

**Reprints and permissions information** is available at www.nature.com/reprints.